\documentclass{article}

\PassOptionsToPackage{numbers, compress}{natbib}

\usepackage[main, final]{neurips_2026}

\usepackage[utf8]{inputenc} 
\usepackage[T1]{fontenc}    
\usepackage{hyperref}       
\usepackage{url}            
\usepackage{booktabs}       
\usepackage{amsfonts}       
\usepackage{nicefrac}       
\usepackage{microtype}      
\usepackage{xcolor}         
\usepackage{amsmath}

\usepackage{multirow}
\usepackage{caption}
\usepackage{array}
\usepackage{listings}
\usepackage{inconsolata}
\usepackage{graphicx}
\usepackage{xcolor}         

\newcommand{\fref}[1]{Fig.~\ref{#1}}
\newcommand{\sref}[1]{Section~\ref{#1}}
\newcommand{\cref}[1]{Chapter~\ref{#1}}

\newcommand{\tref}[1]{Table~\ref{#1}}
\newcommand{\eref}[1]{Eq.~\ref{#1}}
\newcommand{\aref}[1]{Algorithm~\ref{#1}}
\newcommand{\appref}[1]{Appendix~\ref{#1}}

\usepackage{pifont}
\newcommand{\cmark}{\ding{51}}%
\newcommand{\xmark}{\ding{55}}%
\definecolor{greyblue}{RGB}{90,115,200}
\usepackage[table]{xcolor}
\definecolor{lightgreen}{RGB}{220,245,225}
\definecolor{lightred}{RGB}{250,220,220}
\newcommand{\greyblue}[1]{\textcolor{greyblue}{#1}}
\usepackage{tikz}

\usepackage{siunitx}
\usepackage{collcell}

\usepackage{fp}

\usepackage{xcolor}

\newcommand{\layergrid}[1]{%
\tikz[baseline={([yshift=-0.11cm]current bounding box.center)}]{
  \foreach \i in {1,...,12} {
    \draw (\i*0.16,0) rectangle +(0.16,0.22);
  }
  \foreach \j in {#1} {
    \fill[black!50] (0.02+\j*0.16,0.02) rectangle +(0.12,0.18);
  }
}}

\usepackage{subcaption}

\newif\ifstackedtables
\stackedtablestrue

\title{DC-ViT: Modulating Spatial and Channel Interactions for Multi-Channel Images}

\author{
Umar Marikkar$^{1}$ \quad 
Syed Sameed Husain$^{2}$ \quad 
Muhammad Awais$^{1,2}$ \quad 
Sara Atito$^{1,2}$  \\ 
$^1$Surrey Institute for People-Centred AI \hspace{1mm}
$^2$CVSSP \\
University of Surrey, UK\\
{\tt\small \{u.marikkar, sameed.husain, muhammad.awais, sara.atito\}@surrey.ac.uk}
}

\begin{document}

\maketitle

\begin{abstract}
  Training and evaluation in multi-channel imaging (MCI) remains challenging due to heterogeneous channel configurations arising from varying staining protocols, sensor types, and acquisition settings. This heterogeneity limits the applicability of fixed-channel encoders commonly used in general computer vision. Recent Multi-Channel Vision Transformers (MC-ViTs) address this by enabling flexible channel inputs, typically by jointly encoding patch tokens from all channels within a unified attention space. However, unrestricted token interactions across channels can lead to feature dilution, reducing the ability to preserve channel-specific semantics that are critical in MCI data.
To address this, we propose Decoupled Vision Transformer (DC-ViT), which explicitly regulates information sharing using Decoupled Self-Attention (DSA), which decomposes token updates into two complementary pathways: spatial updates that model intra-channel structure, and channel-wise updates that adaptively integrate cross-channel information. This decoupling mitigates informational collapse while allowing selective inter-channel interaction. To further exploit these enhanced channel-specific representations, we introduce Decoupled Aggregation (DAG), which allows the model to learn task-specific channel importances. Extensive experiments across three MCI benchmarks demonstrate consistent improvements over existing MC-ViT approaches. Code is available at \href{https://github.com/umarikkar/DCViT}{https://github.com/umarikkar/DCViT}.
\end{abstract}

\section{Introduction}

Training and evaluating vision encoders on Multi-Channel Imaging (MCI) data is fundamentally challenging due to variability in channel configurations across datasets. Unlike natural images, where three RGB channels are standardised, multi-channel images may contain an arbitrary number of channels whose semantics depend on the data acquisition process.
This variability arises in diverse settings such as multi-channel microscopy, where different staining protocols produce distinct biological markers, and remote sensing, where sensor modalities and fusion strategies determine spectral composition. 
For instance, HPA \cite{thul2017subcellular} contains four channels, JUMP-CP \cite{chandrasekaran2023jump} contains five, WTC-11 \cite{viana2023integrated} contains three, and So2Sat \cite{so2sat} includes 18 channels over multiple spectral bands. This heterogeneity makes it difficult to directly apply fixed-channel encoder designs that are standard in natural image modelling. 

To accommodate variable channel configurations, Multi-Channel Vision Transformers (MC-ViTs) have emerged as the dominant architecture to train on multi-channel datasets \cite{bourriez2024chada,bao2023channel,pham2024enhancing,lian2025isolated}. These models are designed to be channel-adaptive, enabling them to process images with a flexible number of input channels within a unified framework. Following the Vision Transformer (ViT) paradigm \cite{dosovitskiy2020image}, each channel is independently partitioned into respective patch tokens, after which all tokens are concatenated into a single sequence and processed jointly by shared self-attention layers. This long sequence formulation allows information to be exchanged freely across spatial locations and channels throughout the encoder architecture. 

Under this long-sequence formulation, prior work explores mechanisms to mitigate feature homogenisation across channels. This is achieved either by introducing channel-specific tokens \cite{bourriez2024chada,bao2023channel}, or token diversification losses \cite{pham2024enhancing}, or a combination of both, and demonstrates that this increased inter-channel diversity generally improves downstream performance.  However, despite these modifications, the underlying attention mechanism remains fully shared. Consequently, spatial and channel-wise interactions are entangled, limiting explicit control over how cross-channel information is incorporated. 

This unstructured token mixing dilutes channel-specific information within the encoder, thereby limiting the representation capability of MC-ViTs as MCI channels often encode distinct concepts. Yet, complete isolation is also undesirable as certain channels benefit from complementary contextual signals provided by others \cite{gupta2024subcell, c3r}. The objective is thus not to remove cross-channel interaction, but to regulate it. Motivated by this, we propose Decoupled Vision Transformer (DC-ViT), which explicitly regulates inter-channel interactions within the encoder, while maintaining semantic diversity across channels in the learnt representations. DC-ViT is built upon two components, Decoupled Self-Attention (DSA) and Decoupled Aggregation (DAG), which together regulate inter-channel interaction during encoding and representation aggregation.

\begin{figure*}[t]
\centering
\begin{minipage}[c]{0.35\linewidth}
    \centering
    \includegraphics[width=\linewidth]{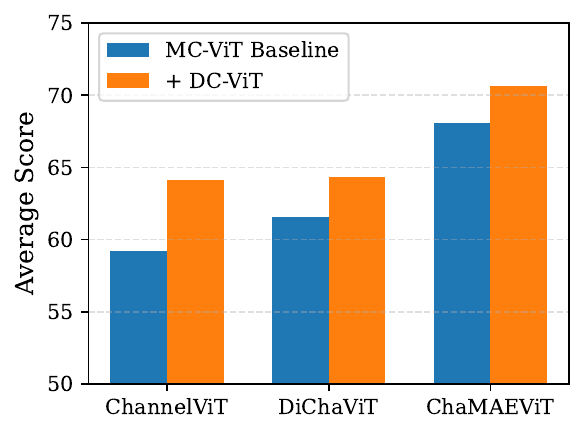}
    \caption{DC-ViT consistently improves existing MC-ViTs across MCI benchmarks. y-axis is the average score over CHAMMI (F1), JUMP-CP (top-1 Acc) and So2Sat (top-1 Acc) downstream tasks.}
    \label{fig:intro}
\end{minipage}%
\hfill
\begin{minipage}[c]{0.6\linewidth}
    \centering
    \includegraphics[width=\linewidth]{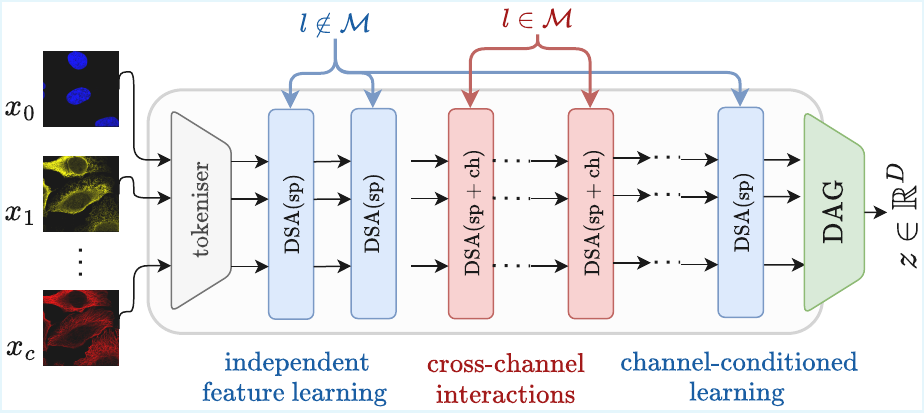}
\caption{Schematic of DC-ViT. Channels build strong low-level representations independently, and are then introduced to cross-channel interactions. DSA(sp) and DSA(sp+ch) denote DC-ViT blocks (\eref{eq:dcvit}) with $l\notin \mathcal{M}$ and $l\in \mathcal{M}$ in \eref{DSA_def}, respectively. }
    \label{fig:arch}
\end{minipage}
\end{figure*}

To regulate inter-channel interactions, we introduce Decoupled Self-Attention (DSA), which factorises token updates into spatial and a channel-wise components. The spatial branch aggregates information across spatial locations within the same channel to reinforce channel-specific features, while the channel-wise branch adaptively integrates information across channels. This separation enables structured cross-channel interaction and reduces the complexity of fully joint attention. To further leverage channel-specific semantics in forming final representations, we propose Decoupled Aggregation (DAG), a hierarchical aggregation scheme that first produces channel-specific embeddings and then aggregates them to form the final representation. This design explicitly models channel contributions to downstream tasks, aligning the representation structure with the heterogeneous nature of MCI datasets and their task formulations.

Together, DC-ViT provides structured regulation of cross-channel information exchange, mitigating feature dilution while retaining beneficial context sharing. As shown in \fref{fig:intro}, across standard MCI benchmarks, DC-ViT consistently outperforms prior MC-ViT baselines.

\section{Background and Related Work}
\label{sec:related}

Our work is motivated by the observation that channels in MCI datasets carry semantically distinct information which are not effectively encoded in existing MC-ViT architectures. Therefore, we first review evidence of channel heterogeneity across MCI datasets, and then review existing MC-ViT architectures.

\paragraph{Channel Heterogeneity in MCI datasets.}  The semantic structure of channels in MCI datasets differs fundamentally from that of RGB images, where channels carry largely redundant information. For example, in immunofluorescent (IF) microscopy, individual channels capture a specific aspect of cellular organisation or phenotype. Some channels mainly provide stable structural context, while others encode signals that vary with the biological condition of interest \cite{c3r}. This semantic variation between channels is present in canonical IF datasets such as JUMP-CP \cite{chandrasekaran2023jump}, HPA \cite{thul2017subcellular}, WTC-11 \cite{viana2023integrated} and Bridge2AI \cite{clark2024cell}. Similarly, semantic variation between channels is observed in satellite-based multi-channel imaging. In the So2Sat LCZ42 dataset, channels contain visible, near-infrared, and short-wave infrared bands, each encoding different physical properties of the Earth’s surface~\cite{so2sat}. Some bands primarily capture stable geometric or reflectance information, while others are more sensitive to surface materials or human activity. As in IF microscopy, these spectral channels are not interchangeable, and modelling approaches must account for their distinct roles.

\paragraph{Multi-Channel Vision Transformers.} Vision encoders designed for multi-channel imaging (MCI) data have been explored across application domains including climate modelling \cite{9672063,nguyen2023climax}, cell microscopy \cite{chen2023chammi,gupta2024subcell,doron2023unbiased}, and satellite imagery \cite{so2sat}. Recent work in this area largely falls under the class of multi-channel Vision Transformers (MC-ViTs), which extend standard Vision Transformers (ViTs) \cite{dosovitskiy2020image} to accommodate MCI data. Unlike vanilla ViTs, which embed spatial patches from all channels into a single token sequence, MC-ViTs tokenise spatial patches from each channel independently, producing channel-specific token embeddings with spatially aligned positional encodings \cite{nguyen2023climax}. Subsequent methods build on MC-ViTs to encourage inter-channel feature diversity by introducing different training strategies or architectural components \cite{bourriez2024chada, bao2023channel, pham2024enhancing, PhamChaMAE2025}. Yet, these methods share a common encoding protocol in which tokens from all channels are concatenated into a single long sequence and processed jointly through self-attention. In contrast, we propose Decoupled Self-Attention, which explicitly separates intra-and-inter channel interactions. 

\paragraph{Decoupling attention between axes.}
The idea of decoupling attention between axis is introduced in Axial attention \cite{ho2019axial}, which computes attention along each axis of a tensor separately and treats the axes symmetrically. ViViT \cite{arnab2021vivit} applies this to the spatial and temporal axes of video with three variants. The factorised encoder stacks a spatial transformer and a temporal transformer, which fuses temporal information late. Factorised self-attention applies spatial and then temporal attention in every block, each with its own parameters. Factorised dot-product attention splits the heads between the two axes, uses separate projections for each, and concatenates the outputs. However, none of these has a dedicated mechanism for setting how much information is shared along the second axis.

DSA differs in two ways. It uses shared query, key, value and output projections for both axes, and it combines the spatial and channel outputs with a learnable scalar $\alpha$ at each channel-attention layer, with channel attention applied only at a subset of layers $\mathcal{M}$. This follows from how MCI differs from video. Adjacent frames are redundant at every depth, so sharing along time is useful throughout the network. In contrast, MCI channels carry distinct information like modalities while remaining spatially aligned, and the useful amount of cross-channel sharing depends on the dataset (\appref{app:results}). Channels also do not yet form useful representations in early layers (\tref{tab:alpha_layers}). 

\section{Methodology}

This section presents the proposed Decoupled Vision Transformer (DC-ViT). We first describe the encoding protocol used in existing multi-channel Vision Transformers (MC-ViTs) in \sref{sec:prelem}. We then introduce DC-ViT and detail its architectural components, including Decoupled Self-Attention and hierarchical pooling for multi-channel data in \sref{sec:dcvit}.

\subsection{Preliminaries}
\label{sec:prelem}

Here, we describe the encoding protocol of existing MC-ViTs. We then examine feature dilution under joint spatial-channel encoding in MC-ViTs and discuss its effect on channel-specific representations.

\paragraph{MC-ViT encoding protocol.}
Let an input MCI image be $X \in \mathbb{R}^{C \times H \times W}$, where $C$ denotes the number of channels and $H, W$ the spatial dimensions. Unlike traditional ViTs, MC-ViTs apply a shared patch embedding layer independently to each channel, producing a representation of shape $C \times N \times D$, where $N$ is the number of tokens per channel and $D$ the feature dimensionality. Channel and spatial positional embeddings are then added, and the representation is flattened along the channel dimension to yield $\mathbf{x}_l \in \mathbb{R}^{CN \times D}$. Each encoder layer then applies the standard ViT update \cite{dosovitskiy2020image} which is a multi-head self-attention (MSA) layer followed by a multi-layer perceptron (MLP) over the full sequence of length $CN$, as illustrated in \fref{fig:mcvit}. Here, the softmax operation in the attention computation is applied across the entire $CN$-length sequence, allowing all spatial tokens across all channels to attend to one another jointly. The detailed description of MC-ViT encoding can be found in \appref{app:architecture}.

\paragraph{Feature dilution in MC-ViTs.} Since attention is performed over the entire $CN$ sequence, there is no explicit constraint on token interactions: any token from any channel and spatial location may aggregate information from all other tokens. This results in reduced feature inter-channel feature diversity between channel-wise representations. To verify this, for each of the MCI datasets, we obtain channel-wise representations under two cases: (1) MC-ViT- where attention is computed over the $CN$ sequence, and (2) Independent attention- where attention is computed individually over each $N$-length sequence for each of the $C$ channels. We then compute the feature cosine similarity between individual channel representations. Over all tested datasets, we find lower cosine similarity between channel representations (0.49 vs.\ 0.72 for WTC-11, 0.48 vs.\ 0.67 for HPA, 0.65 vs.\ 0.76 for JUMP-CP and 0.46 vs.\ 0.53 for So2Sat) when channel attention occurs independently (Case 2) as opposed to jointly (Case 1). Although this is expected as inter-channel interactions are fully restricted in Case 2, the high channel similarity observed from Case 1 may not allow us to leverage the channel heterogeneity in MCI datasets mentioned in \sref{sec:related}.

In contrast, completely avoiding inter-channel interactions is also undesirable, as certain channels provide contextual or reference information that aids the interpretation of others \cite{c3r}. We observe this in preliminary experiments with independently encoded channels (see \tref{tab:alpha_layers}, $\mathcal{M}=\{\}$), where the absence of cross-channel interactions that yield diverse channel representations in fact leads to lower downstream performance. This indicates that while preserving channel-specific features is important, some level of inter-channel interaction remains necessary for MCI tasks, as channels often provide complementary information to one another. Therefore, to preserve feature diversity while enabling effective inter-channel interactions, we propose the Decoupled Vision Transformer (DC-ViT), which controls token interactions explicitly through its architectural design.

\subsection{Decoupled Vision Transformer (DC-ViT)}
\label{sec:dcvit}

\begin{figure*}[t]
\centering
\begin{minipage}[c]{0.27\linewidth}
    \centering
    \includegraphics[height=4.5cm]{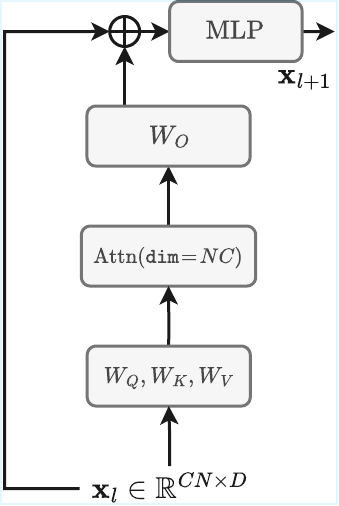}
    \subcaption{}
    \label{fig:mcvit}
\end{minipage}%
\hfill
\begin{minipage}[c]{0.32\linewidth}
    \centering
    \includegraphics[height=4.5cm]{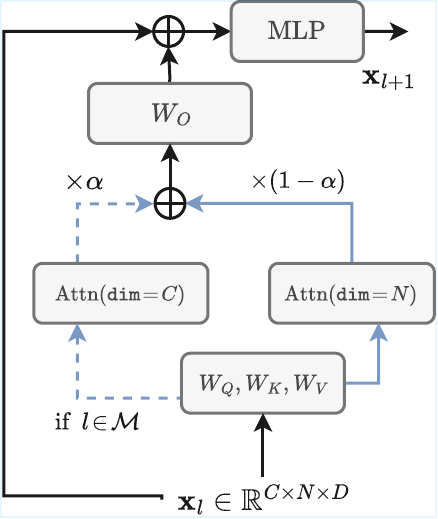}
    \subcaption{}
    \label{fig:dcvit}
\end{minipage}
\hfill
\begin{minipage}[c]{0.32\linewidth}
    \centering
    \includegraphics[height=4.5cm]{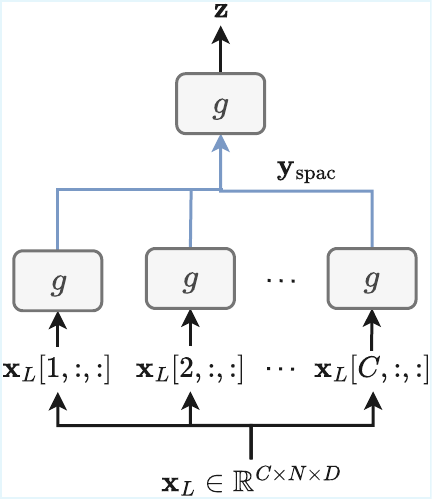}
    \subcaption{}
    \label{fig:DAG}
\end{minipage}
\caption{(a) An MC-ViT layer. Self-attention is performed on a joint set of tokens from all channels. (b) A DC-ViT layer: Self-attention is performed across the set of tokens for a given channel, and across the set of tokens for a given spatial location. Channel attention (dashed path) is only performed when $l \in \mathcal{M}$, in which case $\alpha$ is a learnable scalar. If not, $\alpha$ is fixed to 0. (c) Decoupled Aggregation: Features are aggregated in a hierarchical setting; first spatially and then channel-wise. \greyblue{ $\rightarrow$}: tensor reshaping operations. }
\end{figure*}

Motivated by the observation that channels in MCI datasets encode semantically distinct information, while still being able to  benefit from selective inter-channel information sharing, we introduce the Decoupled Vision Transformer (DC-ViT, \fref{fig:dcvit}). DC-ViT is designed to preserve rich channel-specific representations while enabling cross-channel interactions beneficial for downstream tasks.

DC-ViT consists of two main components. \textbf{Decoupled Self-Attention} decomposes the $NC$-sequence self-attention into spatial ($N$) and channel-wise ($C$) interactions, explicitly separating within-channel and cross-channel token interactions. \textbf{Decoupled Aggregation} aggregates features in a hierarchical manner, first within channels and then across channels, rather than from a unified pool of tokens. We describe Decoupled Self-Attention and Decoupled Aggregation in the following subsections.

\paragraph{Decoupled Self-Attention.}  Let us re-define the input set of patch embeddings as $\mathbf{x} \in \mathbb{R}^{C \times N \times D}$ where $C$ denotes the number of channels, $N$ is the number of tokens per channel, and $D$ is the feature dimensionality. Unlike traditional MC-ViTs, the set of embeddings $\mathbf{x}$ is not flattened to a single channel-sequence dimension (i.e.: $ {C \times N} \rightarrow {CN}$).

At depth $l$ of an encoder of ${L}$ layers, a DC-ViT block computes the output representation $\mathbf{x}_{l+1}$ from the input $\mathbf{x}_l$ by decomposing self-attention into channel-wise and spatial components. Queries, keys, and values are obtained through shared linear projections and are used to compute attention independently across channels and across spatial locations. The resulting attention outputs are combined using a learnable mixing weight $\alpha_l$, and the aggregated representation is then passed through shared output projections and an MLP, following the standard ViT design.

Channel attention is only applied at a subset of encoder layers $\mathcal{M} \subseteq \mathcal{L}$, where $\mathcal{L}$ is the set of encoder layer indices, while spatial attention is applied at all layers. This introduces a channel interaction bottleneck in which the encoder first processes channels independently to capture low-level, channel-specific features, then undergoes selective inter-channel information exchange, and finally encodes  channel-aware  high-level features.

Formally, given the representation $\mathbf{x}_l$ at encoder depth $l$, a DC-ViT block computes the next-layer representation as,
\begin{align}
\label{eq:dcvit}
    \mathbf{x}_{l+1}
    = \mathbf{x}_l
    + \mathrm{MLP}\!\left(
        \mathbf{x}_l + \mathrm{DSA}(\mathbf{x}_l)\right)
    \quad \in \mathbb{R}^{C \times N \times D} .
\end{align}
\noindent
Here, $\mathrm{DSA}(\mathbf{x}_l)$ denotes Decoupled Self-Attention with selective channel interaction, and is defined as,
\begin{align}
\label{DSA_def}
\mathrm{DSA}(\mathbf{x}_l)
=
\begin{cases}
W_O\!\left(
\alpha \cdot \mathrm{Attn}_{ch}(\mathbf{x}_l)
+
(1-\alpha) \cdot \mathrm{Attn}_{sp}(\mathbf{x}_l)
\right),
& \text{if } l \in \mathcal{M}, \\[6pt]
W_O\!\left(
\mathrm{Attn}_{sp}(\mathbf{x}_l)
\right),
& \text{if } l \notin \mathcal{M},
\end{cases}
\end{align}
\noindent
where $W_O$ is the output projection, and $\alpha$ is a learnable scalar for layer $l \in \mathcal{M}$. When $l \notin \mathcal{M}$, channel attention is omitted and the update reduces to a purely spatial attention operation. Unlike the $\mathrm{MSA}$ layer, $\mathrm{DSA}$ explicitly separates channel-wise and spatial token interactions, enabling selective modelling of cross-channel dependencies.

Channel attention shares information across channels at a fixed spatial location and is only applied when $l \in \mathcal{M}$. $\mathrm{Attn}_{ch}$ is defined as,
\begin{align}
\label{ch_attn}
\mathrm{Attn}_{ch}(\mathbf{x}_l)
&=
\bigl[
\mathrm{Attn}\!\left(\mathbf{x}_{l,n}\right)
\bigr]_{n=1}^{N}
\quad \in \mathbb{R}^{C \times N \times D},
\end{align}

\noindent
where $\mathbf{x}_{l,n} \in \mathbb{R}^{C \times D}$ denotes the slice of $\mathbf{x}_l$ at spatial index $n$. Similarly, spatial attention shares information across the $N$ tokens of a given channel and is applied at all layers. $\mathrm{Attn}_{sp}$ is defined as,
\begin{align}
\label{sp_attn}
\mathrm{Attn}_{sp}(\mathbf{x}_l)
&=
\bigl[
\mathrm{Attn}\!\left(\mathbf{x}_{l,c}\right)
\bigr]_{c=1}^{C}
\quad \in \mathbb{R}^{C \times N \times D},
\end{align}

\noindent
where $\mathbf{x}_{l,c} \in \mathbb{R}^{N \times D}$ denotes the token sequence for channel $c$. In both cases, $\mathrm{Attn}(\cdot)$ follows the standard scaled dot-product attention in the original ViT \cite{dosovitskiy2020image}.

\paragraph{Decoupled Aggregation.}  Existing MC-ViTs extract the final representation via the $\mathtt{cls}$ token or global attention pooling, viewing the set of data tokens as a single unified pool, regardless of the assigned channels. Given that via Decoupled Self-Attention, individual channels may exhibit unique semantic concepts that may or may not be dependent on other channels, we implement Decoupled Aggregation (DAG). Here, a pooling operation is sequentially applied within-and-across channels, as opposed to pooling from a unified pool of features. We first define a pooling function,
\begin{align}
    g(\cdot): \mathbb{R}^{N \times D} \mapsto \mathbb{R}^{D},
\end{align}
\noindent which can be any function that reduces the output from a feature map to a single token representation. $g(\cdot)$ may be the extraction of latent tokens such as the $\mathtt{cls}$ token, mean/max pooling, or an attention-based aggregation function \cite{assran2025v,PhamChaMAE2025,carbonneau2018multiple}. 

Given the $L$'th layer representation of the encoder 
$\mathbf{x}_L \in \mathbb{R}^{C \times N \times D}$, Decoupled Aggregation
first performs spatial aggregation within each channel, followed by
aggregation across channels. First, for each channel $c$, we apply a spatial pooling
function $g_{sp}$ over the $N$ tokens,
\begin{align}
\mathbf{y}_{spac, c} &= g_{sp}(\mathbf{x}_{L,c}), \qquad c = 1,\dots,C .
\end{align}

\noindent Stacking the resulting channel representations yields
\begin{align}
\mathbf{y}_{spac} = [\mathbf{y}_{spac, c} ]_{c=1}^{C} \in \mathbb{R}^{C \times D}.
\end{align}

\noindent The final representation is then obtained by
aggregating across channels using the channel pooling function $g_{ch}$,
\begin{align}
\label{eq:DAG}
\mathbf{z} = g_{ch}(\mathbf{y}_{spac}), \qquad \mathbf{z} \in \mathbb{R}^{D}.
\end{align}

Together, DSA and DAG provide an encoding protocol that explicitly controls feature interactions across spatial and channel dimensions. This design allows DC-ViT to maintain channel-specific feature diversity while supporting effective inter-channel information exchange for downstream prediction.

\section{Experiments}

\paragraph{Datasets and Tasks.}  We conduct experiments on three standard benchmarks widely adopted in prior MC-ViT literature \cite{pham2024enhancing, PhamChaMAE2025, lian2025isolated}: CHAMMI, JUMP-CP, and So2Sat. 

 CHAMMI ~\cite{chen2023chammi} is a microscopy benchmark that contains data from WTC-11~\cite{viana2023integrated}, HPA~\cite{hpa-single-cell-image-classification}, and CP~\cite{bray2016cell}. These sources provide images with varying channel counts (3, 4, and 5 fluorescent channels, respectively). The combined benchmark comprises approximately 220K images in total, including 100K training samples. CHAMMI defines nine downstream evaluation tasks, six of which are designed to assess out-of-distribution (OOD) generalisation. In line with existing MC-ViT works, we report the mean OOD performance on the WTC-11 and HPA subsets.

The JUMP-CP dataset~\cite{chandrasekaran2024three} contains multi-channel cellular images acquired under chemical perturbations, with each image consisting of 8 channels (5 fluorescent and 3 bright-field). The dataset is split into 127K training images, 45K validation images, and 45K test images. The downstream task is formulated as a multi-class classification problem with 161 categories, representing 160 distinct compound perturbations and one control condition. 

So2Sat ~\cite{so2sat} is a remote sensing benchmark constructed from satellite imagery collected by Sentinel-1 and Sentinel-2, providing 8 radar channels and 10 multispectral channels, respectively. The dataset includes 352K training samples, along with 24K samples each for validation and testing. The task is posed as a 17-class classification problem, where each class corresponds to a specific land-cover or surface type, such as vegetation, paved regions, or buildings. For both JUMP-CP and So2Sat, we report results under the full channel count during inference, and partial channel counts during inference, following existing works.

\paragraph{Baseline comparisons.}  We evaluate DC-ViT under two settings: (i) the standalone performance of DC-ViT as an encoding protocol, and (ii) its generalisation to pre-trained encoders beyond training from scratch. For architecture comparison, we follow the standard evaluation  in prior MC-ViT studies, where all models are fine-tuned from scratch. To isolate the effect of the encoder architecture, all method-specific training strategies are kept fixed for a fair comparison. For each method, we replace the MC-ViT blocks (\fref{fig:mcvit}) in the encoder backbone of each baseline with the DC-ViT blocks (\fref{fig:dcvit}, \eref{eq:dcvit}), and the pooling function with DAG (\fref{fig:DAG}). 
 
 For supervised baselines, we compare against ChannelViT \cite{bao2023channel}, and DiChaViT \cite{pham2024enhancing}. For hybrid self-supervised + supervised training, we compare with ChaMAEViT \cite{PhamChaMAE2025}. When implementing DC-ViT on the ChaMAEViT baseline, we use the sampling configuration of ChaMAEViT in which the same spatial locations are sampled across channels. The direct comparison with ChaMAEViT under shared masking will isolate the encoder architecture from training strategy. However, this is a limiation of DC-ViT in that 1-1 correspondence is needed between spatial locations across channels for DSA. Nevertheless, we report both sampling settings in \tref{tab:ft_results}.

We additionally evaluate the transfer learning performance of DC-ViT when initialised from pre-trained encoders and adapted to downstream MCI tasks. Experiments are conducted on microscopy datasets, where we pre-train two encoders using iBOT \cite{zhou2021ibot} on HPAv23 \cite{hpa-single-cell-image-classification} and JUMP-CP \cite{chandrasekaran2023jump} single-channel single-cell images, respectively. During pre-training, the encoder learns independent representations for each channel. We perform single-channel pre-training as existing studies show that this transfers effectively to MCI downstream tasks, particularly when the pre-training and fine-tuning channel configurations differ \cite{lian2025isolated, lorenci2025scaling}. For fine-tuning, we adopt the same configuration as ChannelViT, while replacing the attention computation with DSA and aggregation with DAG. 

\paragraph{Training and evaluation.}  Following existing work, we perform our experiments on ViT-S, with patch size 16 for CHAMMI and JUMP-CP, and patch size 8 for So2Sat. Closely following baselines, we set $lr\!=\!4e^{-4}$ for JUMP-CP and So2Sat. For CHAMMI training, we set $lr\!=\!4e^{-4}$ with ChaMAEViT and  $lr\!=\!4e^{-5}$ for others. We set the effective batch sizes (over multi-GPU training) as 128 for JUMP-CP and So2Sat, and 64 for CHAMMI. The remaining hyper-parameters (e.g.: auxiliary loss weights, proxy temperatures) are identical to baselines. In all cases, linear $lr$ scaling is applied where available GPU memory bandwidth is insufficient. All experiments were performed with RTX 3090 or RTX 2080Ti. 

\paragraph{DC-ViT settings.} 
For DSA, we set the channel-attention layer indices from \eref{ch_attn} as $\mathcal{M}=\{4,6,8\}$ for all primary experiments, and the effect of different layer indices for $\mathcal{M}$ can be found in \tref{tab:alpha_layers}. We initialise the scalar channel-interaction weight to $\alpha = 0.1$ across all datasets. Although certain initialisations of $\alpha$ yield stronger metrics, we avoid tuning this parameter to prevent dependence on dataset-specific initialisation, unlike existing works that tune method-specific parameters per dataset \cite{PhamChaMAE2025, pham2024enhancing}. For DAG, We set $g_{sp}$ to be an Attention-Based Multiple Instance Learning  (ABMIL) aggregator, which dynamically assigns a scalar weighting for each feature \cite{carbonneau2018multiple}. Channel pooling function $g_{ch}$ is set as the max-pooling operator, over the channel ($C$) dimension. We compare different choices for $g_{sp}$ and $g_{ch}$, and the effect of $\alpha$ initialisation in \appref{app:results}. 

\section{Results}

\subsection{MCI benchmarks}

We evaluate our method in two settings: (1) fine-tuning from scratch compared to existing MC-ViTs, and (2) fine-tuning from pre-trained checkpoints to assess the transfer learning capability of DC-ViT.

\paragraph{Architecture comparison.}  \tref{tab:ft_results} reports downstream benchmarks compared to existing MC-ViTs. DC-ViT improves the mean on all metrics for ChannelViT, DiChaViT and ChaMAEViT, including when self-supervision is used as an auxiliary objective. For ChaMAEViT, the random sampling baseline is stronger than the shared sampling baseline on all metrics. DC-ViT with shared sampling outperforms it on four of five metrics, with JUMP-CP full as the exception.

\paragraph{Generalisation to pre-trained models.}  \tref{tab:ft_results_pretrain} reports transfer learning performance on MCI benchmarks using different pre-trained encoder checkpoints under the ChannelViT training pipeline. DC-ViT consistently improves over vanilla ChannelViT under both pre-trained models, consistent with the gains observed when training from scratch. This indicates that DC-ViT transfers more effectively from pre-trained encoders than existing MC-ViTs. We attribute this to the fact that commonly used pre-trained models for MCI are largely channel-agnostic, as they are trained on single-channel inputs \cite{lian2025isolated}. Introducing restricted inter-channel interactions during fine-tuning enables the model to incorporate cross-channel information while maintaining compatibility with the representations learnt during single-channel pre-training.

\begin{table*}[t]
	\caption{Performance on MCI benchmarks w.r.t encoder architecture. All methods are trained from scratch. CHAMMI reports mean OOD F1 over WTC-11 and HPA. JUMP-CP and So2Sat report top-1 accuracy. DC-ViT rows show mean $\pm$ std over 5 seeds, and baseline rows are values reported in literature. /S and /R denote shared and random patch sampling paradigms under ChaMAEViT, respectively.}
	\label{tab:ft_results}
	\centering
		\resizebox{\linewidth}{!}{
			\setlength{\tabcolsep}{3pt}
			\begin{tabular}{l l c c c c c c}
				\toprule
				\multirow{2}{*}{Method} & \multirow{2}{*}{} & \multirow{2}{*}{DC-ViT}
				& {CHAMMI}
				& \multicolumn{2}{c}{JUMP-CP}
				& \multicolumn{2}{c}{So2Sat} \\
				\cmidrule(lr){4-4}\cmidrule(lr){5-6}\cmidrule(lr){7-8}
				& & & {Avg.} & {Full} & {Partial} & {Full} & {Partial} \\
				\midrule
				\multirow{2}{*}{ChannelViT} &  & \xmark
				& 64.90  & 67.51  & 56.49 & 61.03 & 46.16 \\
				&  & \cmark
				& \colorbox{lightgreen}{71.43 $\pm$ 0.01} & \colorbox{lightgreen}{72.82 $\pm$ 0.42} & \colorbox{lightgreen}{61.69 $\pm$ 0.31} & \colorbox{lightgreen}{65.15 $\pm$ 0.28} & \colorbox{lightgreen}{48.91 $\pm$ 0.38} \\
				\midrule
				\multirow{2}{*}{DiChaViT} & & \xmark
				& 69.68 & 69.19 & 57.98 & 63.36  & 47.76  \\
				&  & \cmark
				& \colorbox{lightgreen}{71.06 $\pm$ 0.66} & \colorbox{lightgreen}{73.12 $\pm$ 0.25} & \colorbox{lightgreen}{62.12 $\pm$ 0.36} & \colorbox{lightgreen}{65.54 $\pm$ 0.42} & 48.27 $\pm$ 0.52 \\
				\midrule
				\multirow{3}{*}{ChaMAEViT} & /S & \xmark
				& 73.13 & 85.23  & 66.85 & 65.14  & 50.01  \\
				& /R & \xmark
				& 74.63 & \colorbox{lightgreen}{90.73}  & 68.05 & 67.44  & 52.11  \\
				& /S & \cmark
				& \colorbox{lightgreen}{77.91 $\pm$ 0.77} & 85.95 $\pm$ 0.94 & \colorbox{lightgreen}{69.99 $\pm$ 0.44} & \colorbox{lightgreen}{68.28 $\pm$ 1.03} & \colorbox{lightgreen}{52.53 $\pm$ 0.78} \\
				\bottomrule
			\end{tabular}
		}
\end{table*}

\ifstackedtables
\begin{table*}[t]
\centering
\caption{Transfer learning comparison on microscopy benchmarks, applied to the ChannelViT pipeline. Scratch, HPA and JUMP denote the initialisation of the encoder.}
\label{tab:ft_results_pretrain}
\resizebox{0.95\linewidth}{!}{
\setlength{\tabcolsep}{4pt}
\begin{tabular}{c c c c c c c c c c}
\toprule
\multirow{3}{*}{DC-ViT} & \multicolumn{3}{c}{Scratch} & \multicolumn{3}{c}{HPA} & \multicolumn{3}{c}{JUMP} \\
\cmidrule(lr){2-4}\cmidrule(lr){5-7}\cmidrule(lr){8-10}
& CHAMMI & \multicolumn{2}{c}{JUMP-CP} & CHAMMI & \multicolumn{2}{c}{JUMP-CP} & CHAMMI & \multicolumn{2}{c}{JUMP-CP} \\
\cmidrule(lr){2-2}\cmidrule(lr){3-4}\cmidrule(lr){5-5}\cmidrule(lr){6-7}\cmidrule(lr){8-8}\cmidrule(lr){9-10}
& Avg. & Full & Partial & Avg. & Full & Partial & Avg. & Full & Partial \\
\midrule
\xmark & 64.90 & 67.51 & 56.49 & 70.22 & 82.24 & 67.55 & 70.01 & 75.84 & 62.95 \\
\cmark & \colorbox{lightgreen}{71.72} & \colorbox{lightgreen}{73.26} & \colorbox{lightgreen}{61.90} & \colorbox{lightgreen}{77.15} & \colorbox{lightgreen}{83.08} & \colorbox{lightgreen}{68.77} & \colorbox{lightgreen}{73.29} & \colorbox{lightgreen}{78.72} & \colorbox{lightgreen}{65.20} \\
\bottomrule
\end{tabular}
}
\end{table*}
\fi

\subsection{Analysis}

We analyse the contribution of each component and the sensitivity to its configuration. Specifically, we first examine the contribution of DSA and DAG through component ablation, and compare DSA to other attention computations. We then study the placement of inter-channel interaction blocks ($l \in \mathcal{M}$) within the encoder, and finally discuss the computational efficiency of DC-ViT vs.\ the exisitng MC-ViT encoding protocol.

\paragraph{Contribution of DSA and DAG.}  We ablate the two main components of DC-ViT, Decoupled Self-Attention (DSA) and Decoupled Aggregation (DAG). We first replace DSA with joint self-attention in MC-VITs. Under this setting, DAG is carried out by splitting feature tokens of individual channels (i.e.: $NC \rightarrow N \times C$). Secondly, we keep DSA, but replace DAG with joint pooling of all features (i.e.: pooling directly from $NC$ tokens) using the same pooling function. We finally ablate both components, which is equivalent to an MC-ViT (ChannelViT) with attention pooling as opposed to extracting the $\mathtt{cls}$ token. 

From \tref{tab:ablation}, we observe that removing either DSA or DAG from the baseline configuration degrades downstream performance, with a larger drop observed when ablating DSA. When both components are removed, performance deteriorates substantially and is similar to ChannelViT, indicating that both components are necessary for strong performance.

\ifstackedtables\else
\begin{table}[t]
\begin{minipage}{0.41\textwidth}
\centering
\captionof{table}{Ablation of DSA and DAG. Absence of both components is equivalent to ChannelViT with attention pooling.}
\label{tab:ablation}
\resizebox{\linewidth}{!}{
\begin{tabular}{c@{\kern2pt}c@{\kern2pt} c c c}
\toprule
\multicolumn{2}{c}{Components} & CHAMMI & \multicolumn{2}{c}{JUMP-CP} \\
\cmidrule(lr){1-2}\cmidrule(lr){3-3}\cmidrule(lr){4-5}
DSA & DAG & Avg. & Full & Partial \\
\midrule
\cmark & \cmark & \colorbox{lightgreen}{71.72} & \colorbox{lightgreen}{73.26} & 61.90 \\
\cmark & \xmark & 70.02 & 70.70 & \colorbox{lightgreen}{62.41} \\
\xmark & \cmark & 63.46 & 67.03 & \colorbox{lightred}{56.42} \\
\xmark & \xmark & \colorbox{lightred}{62.91} & \colorbox{lightred}{66.15} & 57.20 \\
\bottomrule
\end{tabular}
}
\end{minipage}
\hfill
\begin{minipage}{0.55\textwidth}
\centering
\captionof{table}{Comparison under different attention computations. Applied to ChannelViT pipeline with DAG.}
\label{tab:attention}
\resizebox{\linewidth}{!}{
\setlength{\tabcolsep}{4pt}
\begin{tabular}{c l c c c c}
\toprule
 \multirow{2}{*}{$\mathcal{M}$} & \multirow{2}{*}{Attn.} & CHAMMI & \multicolumn{2}{c}{JUMP-CP} & \multirow{2}{*}{T (ms)} \\
\cmidrule(lr){3-3}\cmidrule(lr){4-5}
 & & Avg. & Full & Partial \\
\midrule
-- & MSA & \colorbox{lightred}{63.46} & \colorbox{lightred}{67.03} & \colorbox{lightred}{56.42} & 26.38 \\
\cmidrule(lr){1-6}
\multirow{2}{*}{$\mathcal{L}$}
& ViViT & 70.78 & 73.27 & 61.98 & 29.95 \\
& Sp $\rightarrow$ Ch & 65.83 & \colorbox{lightgreen}{74.00} & \colorbox{lightgreen}{62.34} & \colorbox{lightred}{30.34} \\
\cmidrule(lr){1-6}
\multirow{3}{*}{$\{4,6,8\}$}
& ViViT & 69.80 & 72.03 & 61.40 & 21.54 \\
& Sp $\rightarrow$ Ch & 69.10 & 72.52 & 61.51 & 22.23 \\
& DSA & \colorbox{lightgreen}{71.72} & 73.26 & 61.90 & \colorbox{lightgreen}{21.35} \\
\bottomrule
\end{tabular}
}
\end{minipage}
\end{table}
\fi

\paragraph{Comparison between types of attention computations.}
We compare DSA against several attention variants in \tref{tab:attention}: standard 
MSA over the full $CN$ sequence, ViViT's factorised dot-product attention 
(fdpa) \cite{arnab2021vivit}, and a sequential spatial-then-channel attention 
(Sp$\,\rightarrow\,$Ch) using the same attention block for both computations. We evaluate ViViT and Sp$\,\rightarrow\,$Ch in cases where all layers contain channel attention ($\mathcal{M=L}$), and where a subset of layers contain channel attention ($\mathcal{M}=\{4,6,8\}$). We set the aggregator to be DAG for all experiments, and only vary the attention computation for analysis. We also measure the average per-sample time taken per training iteration for JUMP-CP (i.e.: forward + backward pass) for a single GPU.

\ifstackedtables
\begin{table}[t]
\begin{minipage}[t]{0.41\textwidth}
\centering
\captionof{table}{Ablation of DSA and DAG. Absence of both components is equivalent to ChannelViT with attention pooling.}
\label{tab:ablation}
\resizebox{\linewidth}{!}{
\begin{tabular}{c@{\kern2pt}c@{\kern2pt} c c c}
\toprule
\multicolumn{2}{c}{Components} & CHAMMI & \multicolumn{2}{c}{JUMP-CP} \\
\cmidrule(lr){1-2}\cmidrule(lr){3-3}\cmidrule(lr){4-5}
DSA & DAG & Avg. & Full & Partial \\
\midrule
\cmark & \cmark & \colorbox{lightgreen}{71.72} & \colorbox{lightgreen}{73.26} & 61.90 \\
\cmark & \xmark & 70.02 & 70.70 & \colorbox{lightgreen}{62.41} \\
\xmark & \cmark & 63.46 & 67.03 & \colorbox{lightred}{56.42} \\
\xmark & \xmark & \colorbox{lightred}{62.91} & \colorbox{lightred}{66.15} & 57.20 \\
\bottomrule
\end{tabular}
}
\end{minipage}
\hfill
\begin{minipage}[t]{0.55\textwidth}
\centering
\captionof{table}{Comparison under different attention computations. Applied to ChannelViT pipeline with DAG.}
\label{tab:attention}
\resizebox{\linewidth}{!}{
\setlength{\tabcolsep}{4pt}
\begin{tabular}{c l c c c c}
\toprule
 \multirow{2}{*}{$\mathcal{M}$} & \multirow{2}{*}{Attn.} & CHAMMI & \multicolumn{2}{c}{JUMP-CP} & \multirow{2}{*}{T (ms)} \\
\cmidrule(lr){3-3}\cmidrule(lr){4-5}
 & & Avg. & Full & Partial \\
\midrule
-- & MSA & \colorbox{lightred}{63.46} & \colorbox{lightred}{67.03} & \colorbox{lightred}{56.42} & 26.38 \\
\cmidrule(lr){1-6}
\multirow{2}{*}{$\mathcal{L}$}
& ViViT & 70.78 & 73.27 & 61.98 & 29.95 \\
& Sp $\rightarrow$ Ch & 65.83 & \colorbox{lightgreen}{74.00} & \colorbox{lightgreen}{62.34} & \colorbox{lightred}{30.34} \\
\cmidrule(lr){1-6}
\multirow{3}{*}{$\{4,6,8\}$}
& ViViT & 69.80 & 72.03 & 61.40 & 21.54 \\
& Sp $\rightarrow$ Ch & 69.10 & 72.52 & 61.51 & 22.23 \\
& DSA & \colorbox{lightgreen}{71.72} & 73.26 & 61.90 & \colorbox{lightgreen}{21.35} \\
\bottomrule
\end{tabular}
}
\end{minipage}
\end{table}
\fi

Unlike the original ViViT and Sp$\,\rightarrow\,$Ch ($\mathcal{L}$), DSA applies channel attention only at $|\mathcal{M}|$ layers, which is $\approx$30\% faster. With $\mathcal{M=L}$, ViViT and Sp$\,\rightarrow\,$Ch perform well on JUMP-CP but worse on CHAMMI, whereas DSA is consistent across both benchmarks and outperforms them on both at the same compute ($\mathcal{M}=\{4,6,8\}$). The learnable scalar $\alpha$ lets DSA control the degree of cross-channel mixing independently of $|\mathcal{M}|$. On So2Sat, where more mixing is useful and $\alpha$ converges to a higher value, DSA performs similarly to Sp$\,\rightarrow\,$Ch and ViViT (\appref{app:results}). The advantage of DSA is largest on CHAMMI, where less mixing is useful.

\paragraph{Robustness to broken patch correspondence.} Unlike MSA in MC-ViTs, DSA relies on a 1-1 correspondence between channels for a given spatial location . This is due to the channel attention component relying on a fixed spatial location. However, exact alignment between channels may not occur in reality. Therefore, we examine 3 violation scenarios: (1) \textbf{channel mis-registration}: a global rigid shift (dx, dy) per channel, random direction; (2) \textbf{local offsets}: a smooth distortion at a random spatial location within a channel; and (3) \textbf{resolution change}: random downsampling per channel.

Under these violations, we compare two settings: trained models evaluated directly under these violations, and models post-trained with the violations as augmentations. In both, we evaluate DSA against MSA under the ChannelViT baseline, across degrees of violation (pixel shift for mis-registration, kernel radius for local offset, and downsampling ratio). The results are shown in \tref{tab:violations}.

We observe that, without training on violations, ChannelViT with standard MSA is more robust than DCViT to increasing degrees of violation. This is expected, as DSA is trained strictly with patch-level correspondence, while for MSA the correspondence is only lightly enforced via positional embeddings. However, by training with violations as augmentations, both the baseline and DC-ViT are able to recover performance. More importantly, DC-ViT now outperforms the ChannelViT baseline across every type of violation and level of severity. This suggests that robustness to patch-level correspondence violations can be achieved by post-training a base model with expected violations.

\begin{table*}
\centering
\caption{Effect on JUMP-CP performance given each violation scenario, applied at a given degree of violation. PT: post-training with violations as augmentations.}
\label{tab:violations}
\resizebox{\linewidth}{!}{
\setlength{\tabcolsep}{2pt}
\begin{tabular}{c c c @{\hskip 18pt} c c c @{\hskip 18pt} c c c @{\hskip 18pt} c c c}
\toprule
\multirow{2}{*}{PT} & \multirow{2}{*}{DSA} & \multirow{2}{*}{orig.} & \multicolumn{3}{c}{Mis-registration} & \multicolumn{3}{c}{Local offsets} & \multicolumn{3}{c}{Res. change} \\
\cmidrule(lr){4-6}\cmidrule(lr){7-9}\cmidrule(lr){10-12}
 && & 2px & 4px & 8px & 2px & 4px & 8px & 2x & 4x & 8x \\
\midrule
\multirow{2}{*}{\xmark} & \xmark &67.51&	\colorbox{lightgreen}{57.52}	& \colorbox{lightgreen}{46.54}	&\colorbox{lightgreen}{26.19}&  \colorbox{lightgreen}{64.14}&	\colorbox{lightgreen}{59.97}	&\colorbox{lightgreen}{50.67} &64.21&	\colorbox{lightgreen}{55.00}	&\colorbox{lightgreen}{33.39} \\
 & \cmark &73.26	&29.95&	17.46	&06.77 &63.63&	49.23&	32.15 	&\colorbox{lightgreen}{67.49}&	48.75	&32.94 \\
\midrule
\multirow{2}{*}{\cmark} & \xmark &67.51	&63.06&	61.81	&61.01 &63.90	&62.16	&57.98 &64.29&	62.96&	56.27 \\
 & \cmark &73.26	&\colorbox{lightgreen}{68.08}&	\colorbox{lightgreen}{63.72}	&\colorbox{lightgreen}{61.86} &\colorbox{lightgreen}{69.82}	&\colorbox{lightgreen}{67.58}&	\colorbox{lightgreen}{62.91}	& \colorbox{lightgreen}{70.66}	& \colorbox{lightgreen}{68.72}	& \colorbox{lightgreen}{63.74} \\
\bottomrule
\end{tabular}
}
\end{table*}

\begin{figure}[t]
\begin{minipage}[t]{0.60\textwidth}
    \centering
    \captionof{table}{Effect of channel attention layer placement. Results are grouped as: early ($\mathcal{M} \subseteq \{1\text{-}5\}$), middle ($\mathcal{M} \subseteq \{4\text{-}8\}$), and late ($\mathcal{M} \subseteq \{8\text{-}12\}$), with values averaged across configurations within each group. In this table, we experiment with ABMIL for pooling operations for DAG.}
    \label{tab:alpha_layers}
    \resizebox{\textwidth}{!}{
    \setlength{\tabcolsep}{3pt}
\begin{tabular}{l  c  c c  c  c c}
\toprule
\multicolumn{2}{l}{\multirow{2}{*}{Layers ($\mathcal{M}$)}} & CHAMMI & \multicolumn{2}{c}{JUMP-CP} & \multicolumn{2}{c}{So2Sat} \\
\cmidrule(lr){3-3}\cmidrule(lr){4-5}\cmidrule(lr){6-7}
&& Avg. & Full & Partial & Full & Partial \\
\midrule
none & \layergrid{} & 66.15 & 47.72 & 44.65 & 64.07 & 49.27 \\
\cmidrule(lr){1-7}
Early & \layergrid{1,2,3} & 71.06 & 68.33 & 59.18 & 63.66 & 48.37 \\
Mid & \layergrid{6,7,8} & \colorbox{lightgreen}{72.07} & {70.41} & {60.31} & \colorbox{lightgreen}{65.98} & 48.68 \\
Late & \layergrid{10,11,12} & 71.63 & 69.39 & 59.61 & 64.24 & 48.87 \\
\cmidrule(lr){1-7}
$\mathcal{L}$(all) & \layergrid{1,2,3,4,5,6,7,8,9,10,11,12} & 69.49 & \colorbox{lightgreen}{71.36} & \colorbox{lightgreen}{60.92} & 65.24 & 49.15 \\
\bottomrule
\end{tabular}
    }
\end{minipage}
\hfill
\begin{minipage}[t]{0.36\textwidth}
    \vspace{0.1em}  
    \centering
    \includegraphics[width=\linewidth]{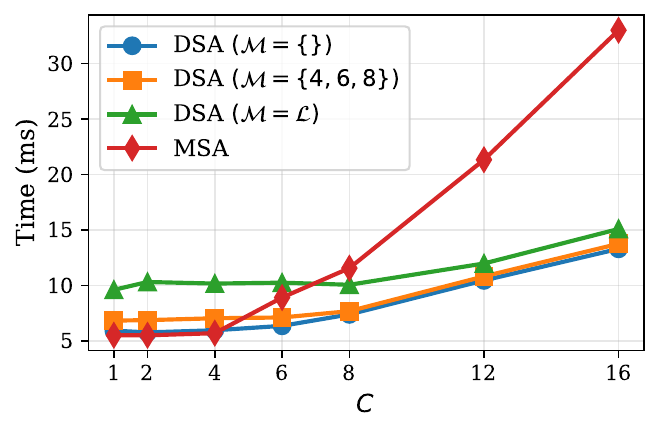}
    \captionof{figure}{Time taken for a single forward pass per sample of DC-ViT vs.\ MC-ViT encoder. For higher channel counts, DSA is much more efficient.}
    \label{fig:timing}
\end{minipage}
\end{figure}

\paragraph{Choice of Channel Attention Layers $\mathcal{M}$.} Restricting channel attention to a subset of layers $\mathcal{M} \subseteq \mathcal{L}$ reduces computational cost (\fref{fig:timing}) and lets the earlier layers build channel-specific representations before any cross-channel exchange. We vary the location of $\mathcal{M}$ with $|\mathcal{M}|=3$, using $\{1,2,3\},\{1,3,5\}$ for Early, $\{5,6,7\},\{4,6,8\}$ for Middle and $\{10,11,12\},\{8,10,12\}$ for Late placement, and report the group averages in \tref{tab:alpha_layers}. Early placement reduces performance, which indicates that early layers benefit from independent low-level feature extraction. Late placement also reduces performance, as DAG favours diverse input features and these are affected by late inter-channel interaction. We therefore use middle layers. The same trend holds for full-channel accuracy on So2Sat (\tref{tab:alpha_layers}).

\paragraph{Computational complexity.} We compare the FLOPs of the attention computation and the runtime of a single forward pass across channel counts, for $\mathcal{M}=\{\}$, $\mathcal{M}=\{4,6,8\}$ and $\mathcal{M}=\mathcal{L}$. For FLOPs we count only the Q-K product, softmax and attention-V product, since projections and MLPs are identical between DSA and MSA. MSA scales as $O(C^2N^2)$ and DSA as $O(CN^2 + NC^2)$, where the $NC^2$ term is negligible for $C \ll N$.

In the runtime measurements (\fref{fig:timing}), DSA has a small overhead at low channel counts and stays stable as $C$ increases, while MSA grows rapidly. The overhead increases with $|\mathcal{M}|$, since layers with channel attention call the attention kernel twice and FlashAttention \cite{dao2022flashattention} calls are not parallelised. We use the $\mathtt{math}$ kernel for channel attention, which leaves a small overhead. For larger channel counts, DSA is significantly faster.

\section{Limitations}
\label{sec:limitations}

The primary limitation of our method is the architectural constraint of requiring a 1-1 mapping between each spatial location across channels. Therefore, implementing random masking algorithms as in ChaMAEViT \cite{PhamChaMAE2025} with DC-ViT is challenging. We aim to address this via methods such as nearest-neighbour or optimal transport alignment to build patch correspondences for inter-channel interactions in future work. The isolated comparison of DC-ViT vs.\ existing MC-ViTs under the ChaMAEViT baseline is with their shared masking setting. Nevertheless, for completeness, we show the comparison with ChaMAEViT random masking in \tref{tab:ft_results}.  Furthermore, there is a lack of MCI datasets for evaluation.  Hence, we are limited to three main benchmarks used in existing MC-ViT works. A broader set of benchmark datasets will help further test the generalisability of DC-ViT and other MC-ViT architectures in general. 

\section{Conclusion}

In this work, we address the challenge of learning effective representations from MCI data, where heterogeneous channel configurations and distinct channel semantics make conventional fixed-channel encoders unsuitable. We introduce Decoupled Vision Transformer (DC-ViT), a framework that explicitly regulates cross-channel information exchange through Decoupled Self-Attention (DSA), and leverages learnt representations effectively for downstream tasks through Decoupled Aggregation (DAG). By separating spatial and channel-wise interactions during encoding and hierarchically aggregating channel representations, DC-ViT preserves channel-specific semantics while enabling beneficial contextual information sharing. Extensive experiments across multiple MCI benchmarks show that incorporating DC-ViT consistently improves the performance of existing MC-ViT architectures. These results indicate that explicitly structuring inter-channel interactions is critical for effectively modelling the heterogeneous and semantically diverse nature of MCI data. Overall, DC-ViT provides a simple yet effective architectural modification that better captures the inherent characteristics of multi-channel imaging datasets, leading to more robust and informative representations for downstream tasks.

\newpage

\bibliographystyle{unsrtnat}
\bibliography{main}


\newpage

\appendix

\section*{Appendix}

\appref{app:architecture} details the MC-ViT encoding protocol for completeness. 
\appref{app:implementation} describes full implementation details, including 
hyperparameter settings and a pseudo-code of DSA. \appref{app:results} presents additional experimental 
results and analysis, including full per-task results on the CHAMMI benchmark, ablations on So2Sat, and ablations 
on the initialisation of $\alpha$ and the choice of pooling functions for 
Decoupled Aggregation (DAG).

\section{MC-ViT architecture}
\label{app:architecture}

Here, we describe in detail the MC-ViT encoding protocol. Let an input MCI image be given as
$X \in \mathbb{R}^{C \times H \times W}$, where $C$ denotes the number of channels and
$H, W$ are the spatial dimensions. 

Unlike traditional ViTs, MC-ViTs apply a shared
patch embedding layer independently to each channel, resulting in an input feature representation
of dimensions ${C \times N \times D}$, where $N$ is the number of tokens per channel
and $D$ is the feature dimensionality. Auxiliary token embeddings (i.e.: channel and spatial positional embeddings) are added to this representation afterwards. 

Existing MC-ViTs flatten this representation along the channel dimension to yield the token representation $\mathbf{x} \in \mathbb{R}^{CN \times D}$, where $CN$ is the resulting sequence length. In practice, the overall sequence length is
$CN + N_{{ext}}$, where $N_{{ext}}$ denotes extra tokens (e.g., the
$\mathtt{cls}$ token); for simplicity, we denote the MC-ViT sequence length as $CN$.

We write the feature update of MC-ViTs as follows, and its schematic can be found in \fref{fig:mcvit}. At encoder depth $l$, the next-layer representation is computed using a standard
ViT update,
\begin{align}
\label{eq:mcvit}
    \mathbf{x}_{l+1}
    = \mathbf{x}_l
    + \mathrm{MLP}\!\left(
        \mathbf{x}_l + \mathrm{MSA}(\mathbf{x}_l)\right)
    \quad \in \mathbb{R}^{CN \times D},
\end{align}
\noindent
where $\text{MLP}_l$ is a two-layer feed-forward network with a $\text{GELU}$ activation. $\mathrm{MSA}_l(\mathbf{x}_l)$ denotes multi-head self-attention applied over the entire sequence as,
\begin{align}
\label{msa}
    \mathrm{MSA}_l(\mathbf{x}_l)
    =
    W_O\!\left(\mathrm{Attn}(\mathbf{x}_l)\right)
    \quad \in \mathbb{R}^{CN \times D},
\end{align}
\noindent
where $W_O$ is the output projection, and $\mathrm{Attn}$ is the computation of scaled dot-product attention, defined as,
\begin{align}
\label{attn}
    \mathrm{Attn}(\mathbf{x}_l)
    =
    \mathrm{Softmax}\!\left(
        \mathbf{q}_l \mathbf{k}_l^{\top} / \sqrt{D}
    \right)\mathbf{v}_l,
\end{align}
\noindent
where
$\mathbf{q}_l = W_Q(\mathbf{x}_l)$,
$\mathbf{k}_l = W_K(\mathbf{x}_l)$, and
$\mathbf{v}_l = W_V(\mathbf{x}_l)$
are obtained from linear projections. When applying \eref{attn} in MC-ViTs, the $\mathrm{Softmax}$ operation is computed
along the full sequence dimension of length $CN$. 

\section{Further implementation details}
\label{app:implementation}

\subsection{Hyperparameter settings}

Code is available at \href{https://github.com/umarikkar/DCViT}{https://github.com/umarikkar/DCViT}.

We use the reported values in existing work for all baseline comparisons \cite{pham2024enhancing, PhamChaMAE2025}. For training, except for the learning rate, we faithfully follow the recommended hyper-parameter settings in existing works \cite{pham2024enhancing,PhamChaMAE2025}. This means, whenever DC-ViT is applied to an existing method, the hyper-parameters of that training setting remain the same as the comparative study, with learning rate adjustments made to validation sets. However, unlike previous works, we perform training with automatic mixed-precision set to FP16 to reduce training time. 

For comparisons against fully supervised baselines (ChannelViT \cite{bao2023channel}, DiChaViT \cite{pham2024enhancing}), we train the models for 60 epochs on CHAMMI, and 100 epochs on JUMP-CP and So2Sat. For CHAMMI training and evaluation, we use the DiChaViT codebase \footnote{\url{https://github.com/chaudatascience/diverse_channel_vit}}, and for JUMP-CP and So2Sat, we use the ChaMAEViT codebase \footnote{\url{https://github.com/chaudatascience/cha_mae_vit}} with the settings [mae\_loss=0, cross\_entropy\_loss=1, decoder=None, depth=12, hcs (ChannelViT), dcs (DiChaViT)]. 

For comparisons against the supervised + self-supervised hybrid ChaMAEViT \cite{PhamChaMAE2025}, we train all models for 100 epochs, and use their codebase for training. However, instead of dynamic patch masking, we perform shared per-channel patch masking where the masking indices are shared throughout the set of sampled channels. 

The learnable scalar weighting $\alpha$ is initialised as an nn.Parameter per layer $l$ where $l \in \mathcal{M}$. For CHAMMI, as training data consists of 3 datasets and these datasets may benefit from different degrees of inter-channel information sharing, we set 3 scalar parameters per layer. We find that this leads to marginally better metrics on CHAMMI evaluation, but a single parameter per layer is sufficient as it still outperforms baselines. Further, a single parameter is simpler to implement as it is dataset agnostic. 

\subsection{DSA pseudo-code}

The Decoupled Self-Attention (DSA) in \aref{lst:dsa}, computes attention independently across two spaces using shared query, key, and value projections. Spatial attention is applied by rearranging the input to treat each channel independently, computing token-to-token interactions across the $N$ patch tokens per channel. For layers $l \in \mathcal{M}$, channel attention is additionally computed by rearranging the input to treat each token independently, capturing channel-to-channel interactions across $C$ channels. The outputs of the two attention computations are then combined via a learnable scalar $\alpha$, and added residually to the input. For all other layers where $l \notin \mathcal{M}$, only spatial attention is applied.

{\scriptsize\ttfamily
\begin{lstlisting}[language=Python, basicstyle=\scriptsize\ttfamily, label=lst:dsa, columns=flexible, keepspaces=true, showstringspaces=false, caption=Pytorch pseudo-code for Decoupled Self-Attention (DSA). $\mathtt{F.sdpa}$ denotes the scaled dot-product attention function in pytorch. Heads are omitted for simplicity.]
class DSA(nn.Module):
  def __init__(self, dim, channel_attn=True, chan_alpha=0.0):
    self.w_q   = nn.Linear(dim, dim)
    self.w_k   = nn.Linear(dim, dim)
    self.w_v   = nn.Linear(dim, dim)
    self.w_o   = nn.Linear(dim, dim)
    if channel_attn:  # l in M
      self.alpha = nn.Parameter(torch.tensor(chan_alpha))
      
  def forward(self, x):  # x: Tensor[B, C, N, D]
    Q, K, V = self.w_q(x), self.w_k(x), self.w_v(x)
    
    with sdpa_kernel(FlashAttention): 
      A_sp = F.sdpa(rearrange(Q,K,V,`b c n d -> (b c) n d'))
    A_sp = rearrange(A_sp, `(b c) n d -> b c n d')
    
    if channel_attn:  # l in M
    
      with sdpa_kernel(Math):  # channel
        A_ch = F.sdpa(rearrange(Q,K,V,`b c n d -> (b n) c d'))
      A_ch = rearrange(A_ch, `(b n) c d -> b c n d')
      A = self.alpha * A_ch + (1 - self.alpha) * A_sp
      
    else:
      A = A_sp
      
    return x + self.w_o(A)
\end{lstlisting}
}

\section{Additional results and analysis}
\label{app:results}

\subsection{Full results on CHAMMI}

\tref{tab:chammi} reports the per-task F1 scores on the CHAMMI benchmark. DC-ViT consistently improves performance across most tasks and methods, with gains observed in the majority of WTC, HPA, and CP dataset tasks, demonstrating the effectiveness of the DC-ViT backbone as a drop-in improvement over standard MC-ViT architectures.

\begin{table}[htbp]
\centering
\caption{F1 Scores of MC-ViT models vs.\ DC-ViT on CHAMMI benchmark. ``OOD'' refers to out-of-distribution tasks. Baseline values are obtained from ChaMAEViT \cite{PhamChaMAE2025}.}
\label{tab:chammi}
\resizebox{\linewidth}{!}{
\setlength{\tabcolsep}{2pt}
\begin{tabular}{l c c c c c c c c c c c c c c}
\toprule
\multirow{2}{*}{Method} & \multirow{2}{*}{DC-ViT} & \multicolumn{4}{c}{Average OOD} & \multicolumn{2}{c}{WTC} & \multicolumn{3}{c}{HPA} & \multicolumn{4}{c}{CP} \\
\cmidrule(lr){3-6}\cmidrule(lr){7-8}\cmidrule(lr){9-11}\cmidrule(lr){12-15}
& & Mean & WTC & HPA & CP & Task1 & Task2 & Task1 & Task2 & Task3 & Task1 & Task2 & Task3 & Task4 \\
\midrule
Channel- & \xmark & 52.54 & 67.58 & 62.35 & \colorbox{lightgreen}{27.81} & 78.36 & 67.58 & 83.93 & 76.73 & 47.97 & 77.70 & \colorbox{lightgreen}{55.16} & 21.89 & \colorbox{lightgreen}{6.38} \\
-ViT& \cmark & \colorbox{lightgreen}{56.93} & \colorbox{lightgreen}{73.04} & \colorbox{lightgreen}{70.40} & 27.35 & \colorbox{lightgreen}{82.00} & \colorbox{lightgreen}{73.04} & \colorbox{lightgreen}{92.01} & \colorbox{lightgreen}{87.92} & \colorbox{lightgreen}{52.87} & \colorbox{lightgreen}{85.95} & 54.31 & \colorbox{lightgreen}{22.28} & 5.45 \\
\midrule
DiCha-   & \xmark & 55.36 & 75.18 & 64.36 & \colorbox{lightgreen}{26.53} & \colorbox{lightgreen}{80.87} & 75.18 & 88.08 & 79.26 & 49.45 & \colorbox{lightgreen}{84.08} & \colorbox{lightgreen}{53.03} & 20.95 & \colorbox{lightgreen}{5.60} \\
-ViT& \cmark & \colorbox{lightgreen}{56.82} & \colorbox{lightgreen}{75.47} & \colorbox{lightgreen}{68.60} & 26.39 & 80.06 & \colorbox{lightgreen}{75.47} & \colorbox{lightgreen}{90.73} & \colorbox{lightgreen}{87.31} & \colorbox{lightgreen}{49.90} & 81.15 & 52.06 & \colorbox{lightgreen}{21.56} & 5.54 \\
\midrule
ChAMAE- & \xmark & 58.02 & 77.15 & \colorbox{lightgreen}{72.11} & \colorbox{lightgreen}{24.81} & 84.52 & 77.15 & 94.14 & 87.47 & \colorbox{lightgreen}{56.75} & 90.89 & \colorbox{lightgreen}{56.68} & \colorbox{lightgreen}{10.25} & 7.50 \\
-ViT& \cmark & \colorbox{lightgreen}{58.54} & \colorbox{lightgreen}{80.38} & 72.37 & 22.87 & \colorbox{lightgreen}{86.29} & \colorbox{lightgreen}{80.38} & \colorbox{lightgreen}{96.25} & \colorbox{lightgreen}{92.46} & 52.28 & \colorbox{lightgreen}{95.38} & 47.65 & 10.10 & \colorbox{lightgreen}{10.86} \\
\bottomrule
\end{tabular}
}
\end{table}

\subsection{Ablations on So2Sat}

We repeat the layer placement and attention comparison experiments on So2Sat with all other settings fixed. \tref{tab:alpha_layers} shows that middle placement gives the best full-channel accuracy (65.98) compared to early (63.66), late (64.24) and no channel attention (64.07). This does not hold for partial-channel accuracy, where the differences are below one point and no placement is clearly best. In \tref{tab:so2sat_attn}, DSA, Sp$\,\rightarrow\,$Ch and ViViT perform similarly with $\mathcal{M}=\{4,6,8\}$. So2Sat results vary more across seeds than CHAMMI and JUMP-CP (\tref{tab:ft_results}), so differences below one point should not be considered conclusive.

\begin{table}[h]
\centering
\caption{Comparison under different attention computations on So2Sat with $\mathcal{M}=\{4,6,8\}$.}
\label{tab:so2sat_attn}
\begin{tabular}{l c c}
\toprule
Attn. & Full & Partial \\
\midrule
Sp$\,\rightarrow\,$Ch & 65.05 & 48.61 \\
ViViT & 65.18 & 48.09 \\
DSA & 65.15 & 48.91 \\
\bottomrule
\end{tabular}
\end{table}

\subsection{Initialisation of $\alpha$.}  We study the effect of the initialisation of $\alpha$ in \tref{tab:alpha_init}. On benchmarks such as CHAMMI, whose downstream tasks rely more heavily on standalone channels, lower initialisation values of $\alpha$ lead to better performance, suggesting that preserving independent channel representations is more beneficial than inducing inter-channel interactions.

However, since $\alpha$ is learnable, it adapts during training according to the task. For instance, when initialised at $0.1$ across all datasets, the final average values (over the set of $\mathcal{M}$ layers) converge to $\alpha=0.107$ for CHAMMI, $\alpha=0.411$ for JUMP-CP, and $\alpha=0.694$ for So2Sat. The initialisation therefore acts mainly as a bias that steers $\alpha$ towards a preferred range and does not require strict dataset-specific tuning. Nevertheless, by initialising $\alpha$ closer to near-optimal values per dataset, given that per-dataset parameter tuning is common practice in existing MC-ViT studies \cite{pham2024enhancing, PhamChaMAE2025}, better performance can be achieved. 

\subsection{Dependence on pooling function.} From \tref{tab:pooling}, we observe that performance remains strong as long as channel importance is learned during the final aggregation stage, on datasets where channel importance is more relevant. This is evident when the $\tt{mean}$ operator is used as the channel aggregator for CHAMMI, where the metrics drop substantially. Using attention-based pooling instead of extracting the per-channel $\tt{cls}$ token yields slightly better results in some cases, but the improvement is not significant. Overall, the key remains adhering to the decoupled aggregation scheme.

\begin{table*}[h]
\centering
\begin{minipage}[t]{0.44\textwidth}
\vspace{2em}
\centering
\captionof{table}{Effect of $\alpha$ value.}
\label{tab:alpha_init}
\resizebox{\textwidth}{!}{
\setlength{\tabcolsep}{1pt}
\begin{tabular}{cccccc}
\toprule
\multirow{2}{*}{$\boldsymbol{\alpha}$}
& CHAMMI & \multicolumn{2}{c}{JUMP-CP} & \multicolumn{2}{c}{So2Sat} \\
\cmidrule(lr){2-2}\cmidrule(lr){3-4}\cmidrule(lr){5-6}
& Avg. & Full & Partial & Full & Partial \\
\midrule
0.00 & \colorbox{lightgreen}{72.09} & 71.80 & 61.23 & 65.16 & 48.46 \\
0.10 & 71.57 & \colorbox{lightred}{71.28} & \colorbox{lightred}{61.13} & \colorbox{lightgreen}{65.76} & \colorbox{lightred}{48.41} \\
0.25 & 71.46 & \colorbox{lightgreen}{72.52} & 61.37 & \colorbox{lightred}{65.10} & 49.05 \\
0.50 & \colorbox{lightred}{69.27} & 71.64 & \colorbox{lightgreen}{61.45} & 65.51 & \colorbox{lightgreen}{49.14} \\
\bottomrule
\end{tabular}
}
\end{minipage}
\begin{minipage}[t]{0.44\textwidth}

\captionof{table}{Effect of pooling function.}
\label{tab:pooling}
\resizebox{0.9\textwidth}{!}{
\setlength{\tabcolsep}{1pt}
\begin{tabular}{clccccc}
\toprule
\multirow{2}{*}{$g_{sp}$} & \multirow{2}{*}{$g_{ch}$} & CHAMMI & \multicolumn{2}{c}{JUMP-CP} \\
\cmidrule(lr){3-3} \cmidrule(lr){4-5}
&& Avg. & Full & Partial \\
\midrule
\multirow{3}{*}{\rotatebox{90}{$\mathtt{cls}$}}
 & $\mathtt{mean}$ & 66.96 & 72.46 & \colorbox{lightred}{60.90} \\
 & $\mathtt{max}$  & 71.14 & 72.34 & 61.35 \\
 & $\mathtt{abmil}$ & \colorbox{lightgreen}{72.09} & 71.56 & 61.44 \\
\cmidrule(lr){1-5}
\multirow{3}{*}{\rotatebox{90}{$\mathtt{abmil}$}}
 & $\mathtt{mean}$ & \colorbox{lightred}{65.43} & 73.01 & 61.52 \\
 & $\mathtt{max}$  & 71.72 & \colorbox{lightgreen}{73.26} & \colorbox{lightgreen}{61.90} \\
 & $\mathtt{abmil}$ & 71.57 & \colorbox{lightred}{71.28} & 61.13 \\
\bottomrule
\end{tabular}
}
\end{minipage}
\end{table*}




\newpage
\section*{NeurIPS Paper Checklist}

The checklist is designed to encourage best practices for responsible machine learning research, addressing issues of reproducibility, transparency, research ethics, and societal impact. Do not remove the checklist: {\bf The papers not including the checklist will be desk rejected.} The checklist should follow the references and follow the (optional) supplemental material.  The checklist does NOT count towards the page
limit. 

Please read the checklist guidelines carefully for information on how to answer these questions. For each question in the checklist:
\begin{itemize}
    \item You should answer \answerYes{}, \answerNo{}, or \answerNA{}.
    \item \answerNA{} means either that the question is Not Applicable for that particular paper or the relevant information is Not Available.
    \item Please provide a short (1--2 sentence) justification right after your answer (even for \answerNA). 
\end{itemize}

{\bf The checklist answers are an integral part of your paper submission.} They are visible to the reviewers, area chairs, senior area chairs, and ethics reviewers. You will also be asked to include it (after eventual revisions) with the final version of your paper, and its final version will be published with the paper.

The reviewers of your paper will be asked to use the checklist as one of the factors in their evaluation. While \answerYes{} is generally preferable to \answerNo{}, it is perfectly acceptable to answer \answerNo{} provided a proper justification is given (e.g., error bars are not reported because it would be too computationally expensive'' or ``we were unable to find the license for the dataset we used''). In general, answering \answerNo{} or \answerNA{} is not grounds for rejection. While the questions are phrased in a binary way, we acknowledge that the true answer is often more nuanced, so please just use your best judgment and write a justification to elaborate. All supporting evidence can appear either in the main paper or the supplemental material, provided in appendix. If you answer \answerYes{} to a question, in the justification please point to the section(s) where related material for the question can be found.

IMPORTANT, please:
\begin{itemize}
    \item {\bf Delete this instruction block, but keep the section heading ``NeurIPS Paper Checklist"},
    \item  {\bf Keep the checklist subsection headings, questions/answers and guidelines below.}
    \item {\bf Do not modify the questions and only use the provided macros for your answers}.
\end{itemize}


\begin{enumerate}

\item {\bf Claims}
    \item[] Question: Do the main claims made in the abstract and introduction accurately reflect the paper's contributions and scope?
    \item[] Answer: \answerYes{} 
    \item[] Justification: The paper mentions the scope (Multi-Channel Imaging) and its contribution (Decoupled Vision Transformer) clearly.
    \item[] Guidelines:
    \begin{itemize}
        \item The answer \answerNA{} means that the abstract and introduction do not include the claims made in the paper.
        \item The abstract and/or introduction should clearly state the claims made, including the contributions made in the paper and important assumptions and limitations. A \answerNo{} or \answerNA{} answer to this question will not be perceived well by the reviewers. 
        \item The claims made should match theoretical and experimental results, and reflect how much the results can be expected to generalize to other settings. 
        \item It is fine to include aspirational goals as motivation as long as it is clear that these goals are not attained by the paper. 
    \end{itemize}

\item {\bf Limitations}
    \item[] Question: Does the paper discuss the limitations of the work performed by the authors?
    \item[] Answer: \answerYes{} 
    \item[] Justification: We discuss limitations in Section~\ref{sec:limitations}.
    \item[] Guidelines:
    \begin{itemize}
        \item The answer \answerNA{} means that the paper has no limitation while the answer \answerNo{} means that the paper has limitations, but those are not discussed in the paper. 
        \item The authors are encouraged to create a separate ``Limitations'' section in their paper.
        \item The paper should point out any strong assumptions and how robust the results are to violations of these assumptions (e.g., independence assumptions, noiseless settings, model well-specification, asymptotic approximations only holding locally). The authors should reflect on how these assumptions might be violated in practice and what the implications would be.
        \item The authors should reflect on the scope of the claims made, e.g., if the approach was only tested on a few datasets or with a few runs. In general, empirical results often depend on implicit assumptions, which should be articulated.
        \item The authors should reflect on the factors that influence the performance of the approach. For example, a facial recognition algorithm may perform poorly when image resolution is low or images are taken in low lighting. Or a speech-to-text system might not be used reliably to provide closed captions for online lectures because it fails to handle technical jargon.
        \item The authors should discuss the computational efficiency of the proposed algorithms and how they scale with dataset size.
        \item If applicable, the authors should discuss possible limitations of their approach to address problems of privacy and fairness.
        \item While the authors might fear that complete honesty about limitations might be used by reviewers as grounds for rejection, a worse outcome might be that reviewers discover limitations that aren't acknowledged in the paper. The authors should use their best judgment and recognize that individual actions in favor of transparency play an important role in developing norms that preserve the integrity of the community. Reviewers will be specifically instructed to not penalize honesty concerning limitations.
    \end{itemize}

\item {\bf Theory assumptions and proofs}
    \item[] Question: For each theoretical result, does the paper provide the full set of assumptions and a complete (and correct) proof?
    \item[] Answer: \answerNA{} 
    \item[] Justification: \answerNA{}
    \item[] Guidelines:
    \begin{itemize}
        \item The answer \answerNA{} means that the paper does not include theoretical results. 
        \item All the theorems, formulas, and proofs in the paper should be numbered and cross-referenced.
        \item All assumptions should be clearly stated or referenced in the statement of any theorems.
        \item The proofs can either appear in the main paper or the supplemental material, but if they appear in the supplemental material, the authors are encouraged to provide a short proof sketch to provide intuition. 
        \item Inversely, any informal proof provided in the core of the paper should be complemented by formal proofs provided in appendix or supplemental material.
        \item Theorems and Lemmas that the proof relies upon should be properly referenced. 
    \end{itemize}

    \item {\bf Experimental result reproducibility}
    \item[] Question: Does the paper fully disclose all the information needed to reproduce the main experimental results of the paper to the extent that it affects the main claims and/or conclusions of the paper (regardless of whether the code and data are provided or not)?
    \item[] Answer: \answerYes{} 
    \item[] Justification: We open-source the code and also provide it in the supplementary material. All implementation details have also been mentioned in the paper itself.
    \item[] Guidelines:
    \begin{itemize}
        \item The answer \answerNA{} means that the paper does not include experiments.
        \item If the paper includes experiments, a \answerNo{} answer to this question will not be perceived well by the reviewers: Making the paper reproducible is important, regardless of whether the code and data are provided or not.
        \item If the contribution is a dataset and\slash or model, the authors should describe the steps taken to make their results reproducible or verifiable. 
        \item Depending on the contribution, reproducibility can be accomplished in various ways. For example, if the contribution is a novel architecture, describing the architecture fully might suffice, or if the contribution is a specific model and empirical evaluation, it may be necessary to either make it possible for others to replicate the model with the same dataset, or provide access to the model. In general. releasing code and data is often one good way to accomplish this, but reproducibility can also be provided via detailed instructions for how to replicate the results, access to a hosted model (e.g., in the case of a large language model), releasing of a model checkpoint, or other means that are appropriate to the research performed.
        \item While NeurIPS does not require releasing code, the conference does require all submissions to provide some reasonable avenue for reproducibility, which may depend on the nature of the contribution. For example
        \begin{enumerate}
            \item If the contribution is primarily a new algorithm, the paper should make it clear how to reproduce that algorithm.
            \item If the contribution is primarily a new model architecture, the paper should describe the architecture clearly and fully.
            \item If the contribution is a new model (e.g., a large language model), then there should either be a way to access this model for reproducing the results or a way to reproduce the model (e.g., with an open-source dataset or instructions for how to construct the dataset).
            \item We recognize that reproducibility may be tricky in some cases, in which case authors are welcome to describe the particular way they provide for reproducibility. In the case of closed-source models, it may be that access to the model is limited in some way (e.g., to registered users), but it should be possible for other researchers to have some path to reproducing or verifying the results.
        \end{enumerate}
    \end{itemize}

\item {\bf Open access to data and code}
    \item[] Question: Does the paper provide open access to the data and code, with sufficient instructions to faithfully reproduce the main experimental results, as described in supplemental material?
    \item[] Answer: \answerYes{} 
    \item[] Justification: Code provided in supplementary material, but needs some cleaning up.
    \item[] Guidelines:
    \begin{itemize}
        \item The answer \answerNA{} means that paper does not include experiments requiring code.
        \item Please see the NeurIPS code and data submission guidelines (\url{https://neurips.cc/public/guides/CodeSubmissionPolicy}) for more details.
        \item While we encourage the release of code and data, we understand that this might not be possible, so \answerNo{} is an acceptable answer. Papers cannot be rejected simply for not including code, unless this is central to the contribution (e.g., for a new open-source benchmark).
        \item The instructions should contain the exact command and environment needed to run to reproduce the results. See the NeurIPS code and data submission guidelines (\url{https://neurips.cc/public/guides/CodeSubmissionPolicy}) for more details.
        \item The authors should provide instructions on data access and preparation, including how to access the raw data, preprocessed data, intermediate data, and generated data, etc.
        \item The authors should provide scripts to reproduce all experimental results for the new proposed method and baselines. If only a subset of experiments are reproducible, they should state which ones are omitted from the script and why.
        \item At submission time, to preserve anonymity, the authors should release anonymized versions (if applicable).
        \item Providing as much information as possible in supplemental material (appended to the paper) is recommended, but including URLs to data and code is permitted.
    \end{itemize}

\item {\bf Experimental setting/details}
    \item[] Question: Does the paper specify all the training and test details (e.g., data splits, hyperparameters, how they were chosen, type of optimizer) necessary to understand the results?
    \item[] Answer: \answerYes{} 
    \item[] Justification: Described in the paper. 
    \item[] Guidelines:
    \begin{itemize}
        \item The answer \answerNA{} means that the paper does not include experiments.
        \item The experimental setting should be presented in the core of the paper to a level of detail that is necessary to appreciate the results and make sense of them.
        \item The full details can be provided either with the code, in appendix, or as supplemental material.
    \end{itemize}

\item {\bf Experiment statistical significance}
    \item[] Question: Does the paper report error bars suitably and correctly defined or other appropriate information about the statistical significance of the experiments?
    \item[] Answer: \answerNo{} 
    \item[] Justification: We follow the default random seeds in existing works. 
    \item[] Guidelines:
    \begin{itemize}
        \item The answer \answerNA{} means that the paper does not include experiments.
        \item The authors should answer \answerYes{} if the results are accompanied by error bars, confidence intervals, or statistical significance tests, at least for the experiments that support the main claims of the paper.
        \item The factors of variability that the error bars are capturing should be clearly stated (for example, train/test split, initialization, random drawing of some parameter, or overall run with given experimental conditions).
        \item The method for calculating the error bars should be explained (closed form formula, call to a library function, bootstrap, etc.)
        \item The assumptions made should be given (e.g., Normally distributed errors).
        \item It should be clear whether the error bar is the standard deviation or the standard error of the mean.
        \item It is OK to report 1-sigma error bars, but one should state it. The authors should preferably report a 2-sigma error bar than state that they have a 96\% CI, if the hypothesis of Normality of errors is not verified.
        \item For asymmetric distributions, the authors should be careful not to show in tables or figures symmetric error bars that would yield results that are out of range (e.g., negative error rates).
        \item If error bars are reported in tables or plots, the authors should explain in the text how they were calculated and reference the corresponding figures or tables in the text.
    \end{itemize}

\item {\bf Experiments compute resources}
    \item[] Question: For each experiment, does the paper provide sufficient information on the computer resources (type of compute workers, memory, time of execution) needed to reproduce the experiments?
    \item[] Answer: \answerYes{} 
    \item[] Justification: Described in the paper. 
    \item[] Guidelines:
    \begin{itemize}
        \item The answer \answerNA{} means that the paper does not include experiments.
        \item The paper should indicate the type of compute workers CPU or GPU, internal cluster, or cloud provider, including relevant memory and storage.
        \item The paper should provide the amount of compute required for each of the individual experimental runs as well as estimate the total compute. 
        \item The paper should disclose whether the full research project required more compute than the experiments reported in the paper (e.g., preliminary or failed experiments that didn't make it into the paper). 
    \end{itemize}
    
\item {\bf Code of ethics}
    \item[] Question: Does the research conducted in the paper conform, in every respect, with the NeurIPS Code of Ethics \url{https://neurips.cc/public/EthicsGuidelines}?
    \item[] Answer: \answerYes{} 
    \item[] Justification: The authors have read the code of ethics.
    \item[] Guidelines:
    \begin{itemize}
        \item The answer \answerNA{} means that the authors have not reviewed the NeurIPS Code of Ethics.
        \item If the authors answer \answerNo, they should explain the special circumstances that require a deviation from the Code of Ethics.
        \item The authors should make sure to preserve anonymity (e.g., if there is a special consideration due to laws or regulations in their jurisdiction).
    \end{itemize}

\item {\bf Broader impacts}
    \item[] Question: Does the paper discuss both potential positive societal impacts and negative societal impacts of the work performed?
    \item[] Answer: \answerNA{} 
    \item[] Justification: \answerNA{}
    \item[] Guidelines:
    \begin{itemize}
        \item The answer \answerNA{} means that there is no societal impact of the work performed.
        \item If the authors answer \answerNA{} or \answerNo, they should explain why their work has no societal impact or why the paper does not address societal impact.
        \item Examples of negative societal impacts include potential malicious or unintended uses (e.g., disinformation, generating fake profiles, surveillance), fairness considerations (e.g., deployment of technologies that could make decisions that unfairly impact specific groups), privacy considerations, and security considerations.
        \item The conference expects that many papers will be foundational research and not tied to particular applications, let alone deployments. However, if there is a direct path to any negative applications, the authors should point it out. For example, it is legitimate to point out that an improvement in the quality of generative models could be used to generate Deepfakes for disinformation. On the other hand, it is not needed to point out that a generic algorithm for optimizing neural networks could enable people to train models that generate Deepfakes faster.
        \item The authors should consider possible harms that could arise when the technology is being used as intended and functioning correctly, harms that could arise when the technology is being used as intended but gives incorrect results, and harms following from (intentional or unintentional) misuse of the technology.
        \item If there are negative societal impacts, the authors could also discuss possible mitigation strategies (e.g., gated release of models, providing defenses in addition to attacks, mechanisms for monitoring misuse, mechanisms to monitor how a system learns from feedback over time, improving the efficiency and accessibility of ML).
    \end{itemize}
    
\item {\bf Safeguards}
    \item[] Question: Does the paper describe safeguards that have been put in place for responsible release of data or models that have a high risk for misuse (e.g., pre-trained language models, image generators, or scraped datasets)?
    \item[] Answer: \answerNA{} 
    \item[] Justification: \answerNA{}
    \item[] Guidelines:
    \begin{itemize}
        \item The answer \answerNA{} means that the paper poses no such risks.
        \item Released models that have a high risk for misuse or dual-use should be released with necessary safeguards to allow for controlled use of the model, for example by requiring that users adhere to usage guidelines or restrictions to access the model or implementing safety filters. 
        \item Datasets that have been scraped from the Internet could pose safety risks. The authors should describe how they avoided releasing unsafe images.
        \item We recognize that providing effective safeguards is challenging, and many papers do not require this, but we encourage authors to take this into account and make a best faith effort.
    \end{itemize}

\item {\bf Licenses for existing assets}
    \item[] Question: Are the creators or original owners of assets (e.g., code, data, models), used in the paper, properly credited and are the license and terms of use explicitly mentioned and properly respected?
    \item[] Answer: \answerYes{} 
    \item[] Justification: Datasets and papers of previous codebases have been cited. 
    \item[] Guidelines:
    \begin{itemize}
        \item The answer \answerNA{} means that the paper does not use existing assets.
        \item The authors should cite the original paper that produced the code package or dataset.
        \item The authors should state which version of the asset is used and, if possible, include a URL.
        \item The name of the license (e.g., CC-BY 4.0) should be included for each asset.
        \item For scraped data from a particular source (e.g., website), the copyright and terms of service of that source should be provided.
        \item If assets are released, the license, copyright information, and terms of use in the package should be provided. For popular datasets, \url{paperswithcode.com/datasets} has curated licenses for some datasets. Their licensing guide can help determine the license of a dataset.
        \item For existing datasets that are re-packaged, both the original license and the license of the derived asset (if it has changed) should be provided.
        \item If this information is not available online, the authors are encouraged to reach out to the asset's creators.
    \end{itemize}

\item {\bf New assets}
    \item[] Question: Are new assets introduced in the paper well documented and is the documentation provided alongside the assets?
    \item[] Answer: \answerNA{} 
    \item[] Justification: \answerNA{}
    \item[] Guidelines:
    \begin{itemize}
        \item The answer \answerNA{} means that the paper does not release new assets.
        \item Researchers should communicate the details of the dataset\slash code\slash model as part of their submissions via structured templates. This includes details about training, license, limitations, etc. 
        \item The paper should discuss whether and how consent was obtained from people whose asset is used.
        \item At submission time, remember to anonymize your assets (if applicable). You can either create an anonymized URL or include an anonymized zip file.
    \end{itemize}

\item {\bf Crowdsourcing and research with human subjects}
    \item[] Question: For crowdsourcing experiments and research with human subjects, does the paper include the full text of instructions given to participants and screenshots, if applicable, as well as details about compensation (if any)? 
    \item[] Answer: \answerNA{} 
    \item[] Justification: \answerNA{}
    \item[] Guidelines:
    \begin{itemize}
        \item The answer \answerNA{} means that the paper does not involve crowdsourcing nor research with human subjects.
        \item Including this information in the supplemental material is fine, but if the main contribution of the paper involves human subjects, then as much detail as possible should be included in the main paper. 
        \item According to the NeurIPS Code of Ethics, workers involved in data collection, curation, or other labor should be paid at least the minimum wage in the country of the data collector. 
    \end{itemize}

\item {\bf Institutional review board (IRB) approvals or equivalent for research with human subjects}
    \item[] Question: Does the paper describe potential risks incurred by study participants, whether such risks were disclosed to the subjects, and whether Institutional Review Board (IRB) approvals (or an equivalent approval/review based on the requirements of your country or institution) were obtained?
    \item[] Answer: \answerNA{} 
    \item[] Justification: \answerNA{}
    \item[] Guidelines:
    \begin{itemize}
        \item The answer \answerNA{} means that the paper does not involve crowdsourcing nor research with human subjects.
        \item Depending on the country in which research is conducted, IRB approval (or equivalent) may be required for any human subjects research. If you obtained IRB approval, you should clearly state this in the paper. 
        \item We recognize that the procedures for this may vary significantly between institutions and locations, and we expect authors to adhere to the NeurIPS Code of Ethics and the guidelines for their institution. 
        \item For initial submissions, do not include any information that would break anonymity (if applicable), such as the institution conducting the review.
    \end{itemize}

\item {\bf Declaration of LLM usage}
    \item[] Question: Does the paper describe the usage of LLMs if it is an important, original, or non-standard component of the core methods in this research? Note that if the LLM is used only for writing, editing, or formatting purposes and does \emph{not} impact the core methodology, scientific rigor, or originality of the research, declaration is not required.
    \item[] Answer: \answerNA{} 
    \item[] Justification: \answerNA{}
    \item[] Guidelines:
    \begin{itemize}
        \item The answer \answerNA{} means that the core method development in this research does not involve LLMs as any important, original, or non-standard components.
        \item Please refer to our LLM policy in the NeurIPS handbook for what should or should not be described.
    \end{itemize}

\end{enumerate}

\end{document}